\documentclass[conference]{IEEEtran}
\IEEEoverridecommandlockouts

\usepackage{cite}
\usepackage{amsmath,amssymb,amsfonts,amsthm,bm,mathtools}
\usepackage{algorithmic}
\usepackage{graphicx}
\usepackage{textcomp}
\usepackage{xcolor}
\usepackage{url}
\usepackage{booktabs}
\usepackage{enumitem}
\usepackage{multirow}
\usepackage{float}
\usepackage{hyperref}

\newcommand{\R}{\mathbb{R}}

\newcommand{\cP}{\mathcal{P}}

\newcommand{\cB}{\mathcal{B}}
\newcommand{\cS}{\mathcal{S}}
\newcommand{\cU}{\mathcal{U}}
\newcommand{\vect}[1]{\boldsymbol{#1}}

\begin{document}

\title{ MultiPathFormer: Towards a Foundation Model for Multipath Wireless Propagation}

\author{
\IEEEauthorblockN{Blessed Guda, Kayley Sze, and Carlee Joe-Wong}
\IEEEauthorblockA{
\textit{Carnegie Mellon University} \\
Pittsburgh, USA \\
blessedg@andrew.cmu.edu, ksze@andrew.cmu.edu, cjoewong@andrew.cmu.edu
}
}

\maketitle

\begin{abstract}
Recent advances in machine learning have enabled training of wireless foundation models, which aim to support tasks such as channel estimation, beam prediction, and localization based on wireless signals.
Existing wireless foundation models typically pretrain on channel tensors using masked reconstruction over subcarriers, antennas, or time but ignore the physical characteristics of wireless propagation. In this work, we propose to instead use multipath propagation as the fundamental pretraining object.
We present MultiPathFormer, an autoregressive foundation model that represents each transmitter--receiver link as an ordered sequence of continuous-valued path tokens and pretrains with next-path prediction. We introduce an Environmental RAG (retrieval augmented generation) mechanism and a first-path codebook on top of the transformer backbone, leveraging environment knowledge to improve path statistics estimation like delay and power by up to 59\%.
MultiPathFormer pretrained on 27 environments transfers to unseen users and, after scenario-specific fine-tuning, outperforms training the corresponding models trained from scratch in new environments. Across downstream tasks, it outperforms SOTA channel-based foundation models, achieving 5.57 m mean localization error, 0.914 top-3 beam accuracy, 0.994 line-of-sight classification accuracy, and 0.561 channel estimation NMSE. These results show that path-level pretraining can learn reusable representations of wireless propagation.
\end{abstract}

 
\section{Introduction}
Modern machine learning applications are often built on top of foundation models, which are pretrained on a general self-supervised objective and then reused across different downstream tasks.  
This approach has produced foundation models with general-purpose representations in domains such as language, vision, robotics and multimodal learning~\cite{bommasani2021opportunities,brown2020language,dosovitskiy2020image,radford2021clip, pmlr-v229-zitkovich23a, reed2022generalistagent}; these representations are then used as the starting point for downstream tasks like text classification, machine translation, generation, and reasoning tasks in natural language processing~\cite{devlin2019bert,vaswani2017attention, guo2025deepseek}.
The shared principle is simple: \textit{with the right pretraining object, large-scale self-supervised learning can learn representations that capture the underlying structure of a domain and are reusable across different tasks.}

The wireless communication domain is another area where foundation models could thrive~\cite{xiao2026wireless}. Emerging  5G/6G wireless communication networks rely on environment-aware inference for tasks like beam selection, channel estimation, localization, and sensing, especially in millimeter-wave and massive multiple-input multiple-output (MIMO) deployments where narrow beams, blockage, and spatial heterogeneity make propagation highly sensitive to the physical scene \cite{Rappaport, 7109864, 3gpp2017study}. At the core of these phenomena is the multipath propagation process. As ElectroMagnetic (EM) waves travel from a transmitter to a receiver, they interact with objects in the environment through reflection, refraction, scattering, and diffraction, generating a collection of propagation paths whose characteristics depend on the underlying scene. While these interactions vary significantly across indoor and outdoor, urban and rural, etc. environments and deployments, 
the propagation process remains governed by the same physical principles. A foundation model that captures them could then be used for downstream tasks.

\textbf{Existing channel-based model architectures.} Several recent works have begun attempts to develop wireless foundation models~\cite{guo2026scalable,guo2026large}. WiFo \cite{liu2025wifo} and ChannelGPT \cite{yu2024channelgpt}, for example, proposed foundation models for channel prediction and showed zero-shot generalisation for the same channel prediction task across heterogeneous Channel State Information (CSI) configurations. The Large Wireless Model \cite{alikhani2024lwm} learns contextualised channel embeddings from large-scale wireless channel data and reports gains on Line of Sight (LoS) classification and beam prediction.  WirelessGPT \cite{yang2025wirelessgpt} extends the same philosophy by using such models on the downstream tasks of human activity recognition and wireless environment reconstruction. The key similarity of all these works is that \textit{they work at the channel level}, with the pretraining objective as some variant of Masked Channel Modelling (MCM)~\cite{jiang2025towards}. The MCM objective is shown in Equation \eqref{eq:mcm}, where the foundation model $f$ makes predictions for all masked channel positions $i \in M$, with the masked set $M$ spanning subcarrier, time, or space and $h_{i}$ denoting the ground truth channel at position $i$.

\begin{equation}
\ell_{MCM} = \frac{1}{|M|} \sum_{i \in M} || f({i|\{h_{j}\}_{j \notin M }\ }) - h_{i} ||^{2}
\label{eq:mcm}
\end{equation}

We ask \emph{whether the channel is the right pretraining object} for wireless foundation models. For a given link, the channel on subcarrier $k$ is a superposition of propagation paths,
\begin{equation}
\vect{H}[k]
=
\sum_{\ell=1}^{L}
\alpha_{\ell}
\exp\!\left(-j2\pi f_k\tau_{\ell}\right)
\vect{a}_{\mathrm{rx}}\!\left(\Omega^{\mathrm{rx}}_{\ell}\right)
\vect{a}_{\mathrm{tx}}^{H}\!\left(\Omega^{\mathrm{tx}}_{\ell}\right),
\label{eq:intro_channel}
\end{equation}
where each path $\ell$ is characterized by a complex gain $\alpha_{\ell}$, delay $\tau_{\ell}$, and arrival/departure angles $\Omega^{\mathrm{rx}}_{\ell}$ and $\Omega^{\mathrm{tx}}_{\ell}$. Equation~\eqref{eq:intro_channel} shows the limitation of channel-level pretraining: the channel tensor mixes a variable number of path events and then further entangles them with array responses, subcarrier sampling, and system configuration. This makes it harder to isolate the environment-dependent structure that actually drives wireless behaviour~\cite{guo2026large}, which may limit the model's interpretability as well as 
its ability to disambiguate the environmental signatures of users at different locations with similar aggregate channels. 

\textbf{Our insight: Path-based architectures.} We argue in this paper that a more fundamental pretraining object is the \emph{multipath process}. Working at the multi path-level exposes the interpretable physical parameters that generate the channel, including path power, delay, phase, angles of arrival and departure, and interaction types such as line-of-sight, reflection, diffraction, and scattering. In addition to interpretability benefits, modelling the multi-path directly may improve reuse of the learned representations, as the model must preserve physical characteristics like variable path cardinality, dominant-path structure, angular geometry, and propagation attributes before they are collapsed into a channel tensor. 
However, this idea raises fundamental \textbf{technical challenges}: multipath foundation modelling does not naturally fit standard masked reconstruction formulations. A transmitter-receiver (tx-rx) link produces a \textit{variable-length sequence} of \textit{continuous-valued path descriptors} rather than a fixed grid of discrete tokens. A useful model must then predict both how many paths exist and each path's physical attributes, while conditioning on the transmitter, receiver, and relevant scene geometry.

We propose \emph{MultiPathFormer}, which formulates wireless pretraining as \emph{next-path prediction}, analogous to \textit{next-token} prediction in language modeling but adapted to continuous wireless path descriptors. Each transmitter-receiver pair is converted into an ordered sequence of propagation-path tokens. An autoregressive transformer is then pretrained over path sequences collected from multiple environments. In designing MultiPathFormer, 
we answer
three fundamental questions:

\begin{enumerate}
\item \textbf{How can a foundation model perform autoregressive prediction over variable-length sequences of continuous-valued path descriptors?} Unlike language tokens drawn from a discrete vocabulary, propagation paths are continuous physical entities characterized by delay, power, phase, and angular parameters, and each tx-rx link may contain a different number of multipaths.
\item \textbf{How should the dominant path be modeled and stabilized during generation?} Unlike text sequences, where tokens at different positions often contribute arbitrarily to the meaning of a sentence, wireless propagation is usually dominated by a single path, and there is no natural temporal path ordering. This dominant path carries a large portion of the received power and largely determines the characteristics of the links. A foundation model must therefore find and treat this path differently from subsequent paths to ensure meaningful, physically consistent generation of subsequent paths. 
\item \textbf{What link-specific environmental context should be retrieved to aid the multipath prediction?} Path formation depends strongly on the surrounding geometry, blockage structures, and potential reflectors, context that is not present in language environments. Knowledge of surrounding objects, e.g., the presence of fixed infrastructure like buildings and transmitter base stations, is often available in practice, e.g., from Open Street Map~\cite{OpenStreetMap}, but a propagation foundation model must determine which are relevant for a particular link and how to encode them.
\end{enumerate}

To the best of our knowledge, \textit{MultiPathFormer is the first wireless foundation model to work at the multipath level}, with a formulation that represents wireless propagation as an autoregressive sequence of individual path tokens and uses \emph{next-path prediction} as the pretraining objective. The \textbf{main contributions} of this paper are summarised below:
\begin{enumerate}
    \item We introduce an \textit{autoregressive wireless path-level formulation} in which the pretraining object is an ordered sequence of multipath tokens rather than the channel tensor. The resulting next-path prediction objective handles continuous-valued, variable-length multipath sequences, and we release MultiPathFormer weights trained on 27 environment scenarios with 23.9M multipath tokens.\footnote{https://huggingface.co/gblessed/multipathformer}
    \item We design an \textit{Environmental RAG mechanism and a KMeans-cluster codebook dominant-path prior} that inject local and corridor-aware scene geometry while stabilizing first-path prediction. Across 31 scenarios, these reduce delay MAE (mean absolute error) by 38.7\% and power MAE by 59.2\% over a direct autoregressive baseline. 
    \item  We show that \textit{pretraining provides both transfer
    and adaptation benefits}: MultiPathFormer
    transfers to unseen users within the pretraining environments, while scenario-specific finetuning for new environments outperforms training the same architecture from scratch.
    \item We evaluate the MultiPathFormer foundation model on \textit{downstream tasks} of beam prediction, user localisation, and LoS classification. Experimental results show that the MultiPathFormer outperforms baseline channel foundation models with improvements of up to 30\% on Beam prediction and 3\% on LoS classification, 65m distance on user localization, and 0.751 Channel estimation NMSE. 

\end{enumerate}

\section{Literature Review}
Broad self-supervised pretraining and downstream task reuse form the basis of foundation models \cite{bommasani2021opportunities}. Token-level objectives can provide useful representations for language tasks, as demonstrated by BERT-style masked-token modeling and GPT-style autoregressive next-token prediction \cite{devlin2019bert,brown2020language}, as well as for image patches and image–text pairs in vision and multimodal learning by Vision Transformers, Masked Autoencoders, and CLIP-style contrastive learning \cite{he2021maskedautoencodersscalablevision, dosovitskiy2020image,  radford2021clip}. MultiPathFormer takes inspiration from these ideas in using propagation-path tokens for wireless foundation modelling.

\subsection{Wireless Foundation Models}
Recent works have proposed foundation models for wireless communications and sensing~\cite{guo2026scalable}, particularly at the channel level as surveyed by~\cite{jiang2025towards}.
LWM \cite{alikhani2024lwm} pretrained a BERT-style encoder-only model with masked channel modeling across antenna and subcarrier dimensions, producing contextualized embeddings useful for downstream tasks such as LoS classification and beam prediction.
WiFo \cite{liu2025wifo} extended this direction by modeling CSI evolution over time with 3D channel inputs, allowing the foundation model to capture temporal, frequency, and spatial variations. WiFo uses a masked autoencoder (MAE) \cite{he2021maskedautoencodersscalablevision} with a transformer decoder to increase reconstruction capacity, which \cite{guler2025robust} extends with a contrastive objective to produce more discriminative representations. Other works have proposed physics-inspired architectures at the channel ~\cite{chen2026spa} or EM~\cite{xiao2026wireless} levels and attempted to exploit the timeseries nature of wireless signals~\cite{sheng2025wireless,fontaine2024towards}, but they do not use path-level tokens.
WirelessGPT \cite{yang2025wirelessgpt} and ChannelGPT \cite{yu2024channelgpt} aim to integrate such wireless foundation models into the broader network, respectively extending them to sensing tasks and feedback loops that allow wireless foundation models to respond to changes in the physical environment and network performance. Other works leverage the availability of multi-modal environment information, integrating it into wireless foundation models~\cite{zhang2025multi, han2025wicomg}, as does MultiPathFormer. 

These works show that self-supervised MCM models can produce reusable representations. All of them, however, \emph{use the channel tensor as the pretraining object}, tying the learned representation to the antenna, subcarrier, and bandwidth configuration and only implicitly capturing 
the multipath events that compose the channel. MultiPathFormer instead directly pretrains on the multipath sequence.







\subsection{Task-Specific Wireless Models}

Before the advent of foundation models, most deep learning models used in communication were trained for a single task. For example, \cite{alrabeiah2020beamblockage} showed that large MLP (multi-layer perceptron) models can take sub-6 GHz channels and predict mmWave beams and blockage states. \cite{8395149} similarly mapped low-overhead pilot observations to beam selections with deep neural networks, while CsiNet~\cite{li2023autocsinetscenariocustomizedautomaticneural} introduced an autoencoder-style neural architecture for massive-MIMO CSI compression and channel reconstruction, 
ChannelNet \cite{soltani2019deeplearningbasedchannelestimation} framed the channel estimation problem as an image super-resolution and restoration problem over the time-frequency grid, and \cite{8353153} used the image denoising formulation and applied a message-passing denoising convolutional neural network for the channel estimation of sparse mmWave massive-MIMO channels. These works show the value of neural networks for channel recovery, but they remain channel-matrix methods 
rather than working on an explicit sequence of propagation paths.
\section{MultiPathFormer Design}

We first introduce our problem formulation (Section~\ref{subsec:problem_formulation}) before presenting the MultiPathFormer backbone architecture in Section~\ref{sub_sec:pathformer_backbone}. We then introduce the environmental RAG (Section~\ref{sub_sec:env_rag}) and cluster codebook (Section~\ref{sub_sec:codebook}) additions that respectively allow MultiPathFormer to incorporate known environment context and stabilize the dominant path.

\subsection{Background and Problem Formulation}\label{subsec:problem_formulation}
\begin{figure*}[t]
    \centering
    \begin{minipage}[t]{0.45\textwidth}
        \centering
        \includegraphics[height=3.0cm, width=\linewidth]{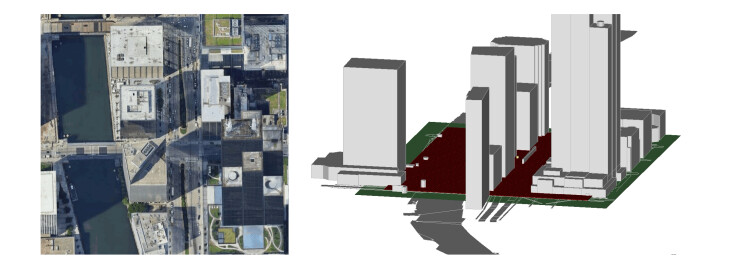}
        
        \vspace{2pt}
        \small \textbf{a)} Physical Environment and 3D user grid
    \end{minipage}
    \hfill
    \begin{minipage}[t]{0.45\textwidth}
        \centering
        \includegraphics[height=3.0cm, width=\linewidth]{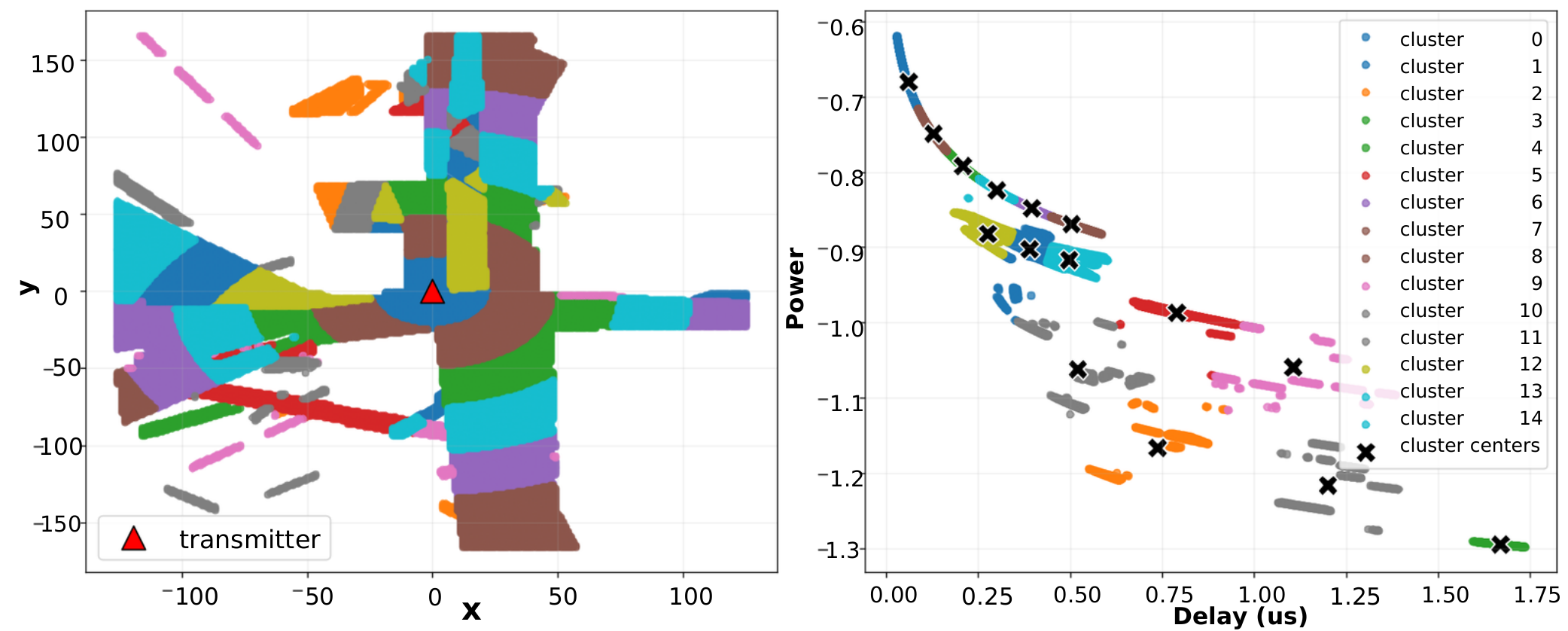}
        
        \vspace{2pt}
        \small \textbf{b)} Physical environment
    \end{minipage}
    \hfill
    \caption{
    \textbf{a)} An example physical environment  and the corresponding 3D model through which wireless signals may propagate.
    \textbf{b)} User clusters obtained from KMeans on first-path delay and power (right), showing that the clusters organize users into spatially coherent groups (left) under a fixed transmitter. 
    }
    \label{fig:env_kmeans_overview}
\end{figure*}

\textbf{Channel model.} We consider a training setup containing $M$ environments; note that MultiPathFormer can later be finetuned for new environments, as we show in Section~\ref{sub_sec:finetune}. For each environment $m \in \cS = \{1,\dots,M\}$, we consider the multipath propagation between transmitters $b \in \cB_m$ and receivers $u \in \cU_m$ distributed throughout the environment.  In an environment $m$, for each tx-rx pair $(m,b,u)$, we get the tx and rx geographical positions $\vect{x}_{\mathrm{tx}}^{(m,b)}, \vect{x}_{\mathrm{rx}}^{(m,u)}  \in \R^3$, link environment features $\vect{x}_{m}^{tx-rx} \in \R^{d_{env}}$, and the corresponding channel tensor $\vect{H}_{m,b,u}$. We show an example of such an environment in Figure \ref{fig:env_kmeans_overview}{a}. The corresponding multipath sequence $\cP_{m,b,u}$
 with $L_{m,b,u}$ number of mulitpath is defined as
\begin{align}
\cP_{m,b,u} &= \big(\boldsymbol{\pi}^{(m,b,u)}_1,\ldots,\boldsymbol{\pi}^{(m,b,u)}_{L_{m,b,u}}\big),
\label{eqn:multpathset} \\
%
\boldsymbol{\pi}^{(m,b,u)}_\ell &= \Big(\rho_\ell,  \tau_\ell,\phi_\ell, \theta^{\mathrm{aoa}}_{\ell,\mathrm{az}}, \theta^{\mathrm{aoa}}_{\ell,\mathrm{el}}, \theta^{\mathrm{aod}}_{\ell,\mathrm{az}}, \theta^{\mathrm{aod}}_{\ell,\mathrm{el}}, \vect{z}_\ell\Big).
\label{eq:path_descriptor}
\end{align}
Each multipath $\boldsymbol{\pi}^{(m,b,u)}_\ell$ component for each path $\ell$ is defined in Equation \eqref{eq:path_descriptor}, where $\rho_\ell$ is the path power, $\tau_\ell$ the propagation delay, $\phi_\ell$ the phase, $\theta^{\mathrm{aoa}}_{\ell,\mathrm{az}}$ and $\theta^{\mathrm{aoa}}_{\ell,\mathrm{el}}$ are the azimuth and elevation angles of arrival, $\theta^{\mathrm{aod}}_{\ell,\mathrm{az}}$ and $\theta^{\mathrm{aod}}_{\ell,\mathrm{el}}$ are the corresponding angles of departure, and $\vect{z}_\ell\in  \{0,1\}^4 $ is 
a multihot vector on interaction type showing if the path experiences Reflection, Diffraction, Scattering, or Line of Sight (LoS).   We omit the $(m,b.u)$ superscript in components of $\boldsymbol{\pi}$ for notational simplicity.
 Equations \eqref{eq:mcm}  - \eqref{eq:path_descriptor} are the standard multipath model as used in other papers \cite{DBLP:journals/corr/abs-1902-06435, liu2025wifo}.

\textbf{Autoregressive path formulation.} For a tx-rx pair and its environmental features, we formulate the autoregressive model in Equation \eqref{eq:next_path} to predict the arrival of the next multipath given the previous multipaths. Here $\theta$ denotes the model parameters and we compute the probability of the sequence $\cP_{m,b,u}$: 
\begin{equation}
p_{\theta}(\cP_{m,b,u}\mid \mathbf{x}_m)
=
\prod_{t=1}^{L_{m,b,u}}
p_{\theta}
(\boldsymbol\pi_t \mid \boldsymbol\pi_{<t},
\mathbf{x}_{\rm rx},
\mathbf{x}_{\rm tx},
\mathbf{x}_m^{tx-rx})
\label{eq:next_path}
\end{equation}

This formulation leads us to \textit{solving an autoregressive variable sequence length regression problem}. This is different from the standard autoregressive formulation in language modelling: first, in language modelling, tokens are derived from the text, yielding a discrete vocabulary, and the model solves a \textit{variable sequence length autoregressive classification problem} over the vocabulary. Secondly, in language modelling, start and stop tokens are added to the vocabulary to initiate and terminate generation, which does not translate to a regression setup. Finally, wireless multipaths may have different orderings, unlike the temporal order of language tokens.

To solve these challenges, we first convert the path set into a deterministic ordered sequence by sorting them by descending power $\rho_{1}^{m,b,u} \ge \rho_{2}^{m,b,u} \ge \cdots \ge  \rho_{L_{m,b,u}}^{m,b,u}$. The other option is to sort based on delay, because; as shown in Figure \ref{fig:env_kmeans_overview}{b}, received power and delay not have a perfect inverse-linear relationship. However, we do not find any significant performance differences between the two setups. We prepend a zero-valued start token $\boldsymbol\pi^{(m,b,u)}_0=\mathbf{0}$ to initiate generation. At each timestep, the model simultaneously predicts the path parameters defined in Equation         \eqref{eq:path_descriptor}.  To terminate generation, rather than define a target end-of-sentence path, which could hurt the performance in a regression setup, we explicitly predict the number of paths between the transmitter and receiver and terminate based on this prediction. Note that, since later paths are expected to have lower power, they likely contribute less to the overall signal, and thus minor inaccuracies in termination likely do not have much effect on our channel prediction ability. 

\begin{figure}
    \centering
    \includegraphics[width=0.98\linewidth]
    {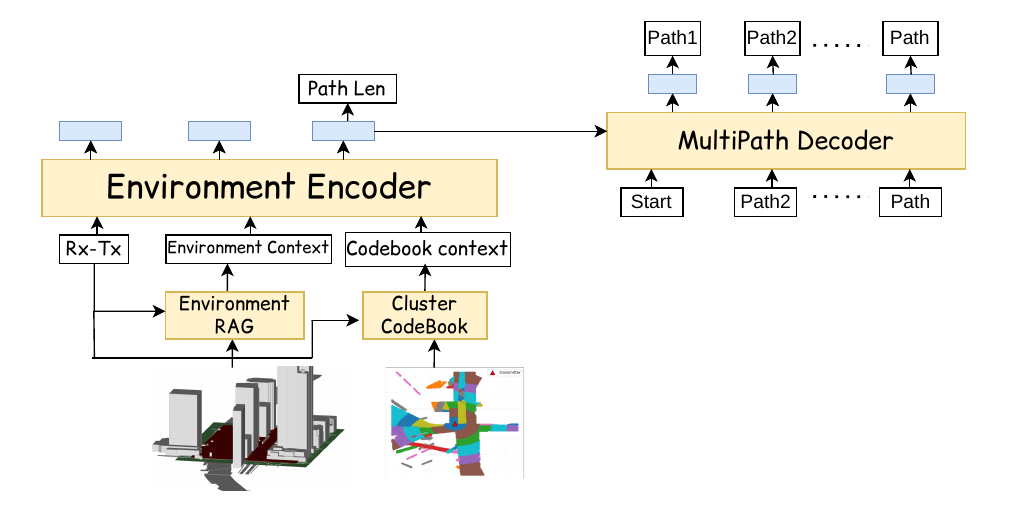}
    \caption{
    Overview of the proposed corridor-aware MultiPathFormer. The model augments the tx-rx prompt with Environmental RAG features and a first-path KMeans codebook, then autoregressively predicts the multipath sequence with a residual correction for the dominant first path.
    }
    \label{fig:pathformer_corridor}
\end{figure}

The remaining challenge is how to \textit{condition} this sequence model on the right physical context. The full environment may contain many objects that are irrelevant to a particular link, while the dominant first path often follows a strong spatial pattern under a fixed transmitter. MultiPathFormer therefore combines three pieces as shown in Figure \ref{fig:pathformer_corridor}: an Environmental RAG module (Section~\ref{sub_sec:env_rag}) that retrieves pair-specific scene geometry and a compact First-Path Cluster Codebook (Section~\ref{sub_sec:codebook}) that supplies a low-dimensional prior for the dominant path supplement a transformer backbone (Section~\ref{sub_sec:pathformer_backbone}). 

\subsection{MultiPathFormer Backbone Model}\label{sub_sec:pathformer_backbone}
The MultiPathFormer backbone uses an encoder-decoder transformer architecture similar to vision-language models \cite{li2023trocr}, with the encoder and decoder capturing information from different modalities. The encoder produces representations capturing the relationship between tx-rx and the physical environment. To encourage the encoder to produce environmentally grounded representations and to address the variable sequence length, we supervise the encoder representation to predict the number of multipaths between the tx-rx pair. The decoder then captures how the current path relates to previous paths conditioned on the environment encoder representation. 
Letting $\vect{E}_{\mathrm{env}}$ denote the encoder output, $\hat{L}_{m,b,u}$ the estimated path length, and $\vect{h}_t$ the decoder output, we summarize MultiPathFormer at generation step $t$ in Equations \eqref{eq:pathformer_encoder}-\eqref{eq:pathformer_decoder}. 
\begin{gather}
\vect{E}_{\mathrm{env}}
=
\mathrm{Encoder}_{\theta}\bigl([\mathbf{x}_{m}, \mathbf{x}_{tx}, \mathbf{x}_{rx}]\bigr),
\label{eq:pathformer_encoder}
\\
\hat{L}_{m,b,u}
=
f_{\mathrm{len}}(\vect{E}_{\mathrm{env}}),
\label{eq:pathformer_length_head}
\\
\vect{h}_t
=
\mathrm{Decoder}_{\theta}\bigl([\vect{E}_{\mathrm{env}};\boldsymbol{\pi}_{<t}]\bigr).
\label{eq:pathformer_decoder}
\end{gather}

We then pass the decoder output through a linear layer to estimate the multipath parameters defined in Equation~\eqref{eq:path_descriptor}:
\begin{align}
&\begin{aligned}
\widehat{\vect{y}}_t
&=
\Bigl(
\hat{\tau}_t, \hat{\rho}_t,
\widehat{\sin\phi_t}, \widehat{\cos\phi_t},
\widehat{\sin\theta^{\mathrm{aoa}}_{t,\mathrm{az}}},
\widehat{\cos\theta^{\mathrm{aoa}}_{t,\mathrm{az}}},
\\
&\qquad
\widehat{\sin\theta^{\mathrm{aoa}}_{t,\mathrm{el}}},
\widehat{\cos\theta^{\mathrm{aoa}}_{t,\mathrm{el}}},
\widehat{\sin\theta^{\mathrm{aod}}_{t,\mathrm{az}}},
\\
&\qquad
\widehat{\cos\theta^{\mathrm{aod}}_{t,\mathrm{az}}},
\widehat{\sin\theta^{\mathrm{aod}}_{t,\mathrm{el}}},
\widehat{\cos\theta^{\mathrm{aod}}_{t,\mathrm{el}}}
\Bigr)
=
W_{\mathrm{reg}}\vect{h}_t + \vect{b}_{\mathrm{reg}}
\end{aligned}
\label{eq:pathformer_reg_head} \\
&\hat{\vect{z}}_t
=
\sigma(\vect{s}_t),\quad \vect{s}_t
=
W_{\mathrm{int}}\vect{h}_t + \vect{b}_{\mathrm{int}}.
\label{eq:pathformer_interaction_head}
\end{align}

\textbf{Training loss.} For a path with length $L_{m,b,u}$, groundtruth path $y_t$ derived from $\pi_\ell^{m,b,u}$, we compute an $\ell_2$ loss $\mathcal{L}_{\mathrm{path}}$ for the path parameters and a squared loss $\mathcal{L}_{\mathrm{len}}$ for the path length. The loss $\mathcal{L}_{\mathrm{int}}$ in estimating $\vect{z}_t$ is defined with cross-entropy loss, as $\vect{z}_t$ is a categorical variable: 
\begin{align*}
\mathcal{L}_{\mathrm{path}}
&=
\frac{1}{L_{m,b,u}}
\sum_{t}
\left\|
\hat{\vect{y}}_t - \vect{y}_t
\right\|_2^2,
\\
\mathcal{L}_{\mathrm{len}}
&=
\left(
\hat{L}_{m,b,u} - L_{m,b,u}
\right)^2,
\\
\mathcal{L}_{\mathrm{int}}
&=
-\frac{1}{L_{m,b,u}}
\sum_{t}
\Bigl[
\vect{z}_t^{\top}\log \hat{\vect{z}}_t
+
(1-\vect{z}_t)^{\top}\log(1-\hat{\vect{z}}_t)
\Bigr]
\end{align*}
We use standard backpropagation methods to train the model parameters $\left(\theta, W_{\mathrm{int}}, \vect{b}_{\mathrm{int}}, W_{\mathrm{reg}}, \vect{b}_{\mathrm{reg}}\right)$ on the total loss function $\mathcal{L} =
 \mathcal{L}_{\mathrm{path}}
 +
 \lambda_{\mathrm{int}}\mathcal{L}_{\mathrm{int}}
 +
 \lambda_{\mathrm{len}}\mathcal{L}_{\mathrm{len}}$ over the training data points $(m,b,u)$.
We  specify the $\lambda_{\mathrm{len}}$ and $\lambda_{\mathrm{int}}$ weights and other configurations in the Appendix.

\subsection{Environmental Retrieval Augmented Generation}
\label{sub_sec:env_rag}

During propagation, multipaths interact with a variety of objects in the environment. Environments can vary starkly, with sparser buildings in less populated areas to more dense objects and users in urban settings. We, however, require our model to be able to reason about relevant objects in all types of environments in an efficient manner.
The na\"ive approach of passing a single global environment descriptor $\vect{x}_m$ may be too coarse and noisy for path prediction, especially when disambiguating users across heterogeneous environments. The paths for a tx-rx pair are determined not only by which environment the pair belongs to, but more specifically by \emph{which objects are locally close to the receiver and which objects lie along the tx-rx propagation corridor}. This motivates our \textit{Environmental Retrieval RAG} framework, which retrieves tx-rx-specific geometric context from the scene and conditions MultiPathFormer on this retrieved representation.

For an environment $m$, with tx-rx pair located at $\vect{x}^{(m,b)}_{\mathrm{tx}}, \vect{x}^{(m,u)}_{\mathrm{rx}} \in \mathbb{R}^2$ , let $\mathcal{O}_m=\{o^{(m)}_1,\dots,o^{(m)}_{J_m}\}$
denote the set of physical objects in the scene. These objects can easily be obtained for any environment through the 3D model of the environment produced from OpenStreetMap \cite{OpenStreetMap} and are also used in prior work \cite{s26072223}.  Each object $o^{(m)}_j$ is associated with geometric attributes such as planar centroid $\vect{p}^{(m)}_j\in\mathbb{R}^2$, height $h^{(m)}_j$, footprint area $a^{(m)}_j$, and material properties.
Our Environmental RAG module retrieves the local and corridor objects as well as sparse multiscale summaries defined in the subsequent paragraphs.
\paragraph{Local dense retrieval}
For a query point $\vect{q}\in\mathbb{R}^2$, which may be the rx or tx location $\vect{x}^{(m,u)}_{\mathrm{rx}},\vect{x}^{(m,b)}_{\mathrm{tx}}$, we define the local object candidate set within radius $d_{\max}$ in Equation \eqref{eq:candidate_set} and retain the $K$ nearest objects $\mathcal{N}_{K}(\vect{q}, d_{\max})$:
\begin{equation}
\mathcal{O}_{m}(\vect{q}, d_{\max})
=
\left\{
o_j \in \mathcal{O}_m :
\left\|
\vect{p}^{(m)}_j-\vect{q}
\right\|_2 \le d_{\max}
\right\},
\label{eq:candidate_set}
\end{equation}
\begin{equation}
\mathcal{N}_{K}(\vect{q}, d_{\max})
=
\operatorname{TopK}_{o_j \in \mathcal{O}_{m}(\vect{q}, d_{\max})}
-\left(\left\|
\vect{p}^{(m)}_j-\vect{q}
\right\|_2 \right) .
\label{eq:local_retrieval}
\end{equation}
For each object in $\mathcal{N}_{K}(\vect{q}, d_{\max})$, we keep a dense description of distance, height, footprint area, and material properties.

\paragraph{Corridor dense retrieval}
To capture objects that are not close to either endpoint but still lie near the propagation route, we further define a tx-rx corridor. We project the centroid $\vect{p}_{j}^{(m)}$ of each object $o_j$ onto the tx-rx line using Equation \eqref{eq:corridor_segment} and then compute the distance of the centroid to the projected point $\vect{v}_{j}^{m,b,u}$  in Equation \eqref{eq:corridor_distance}. We then retrieve the Top-k close objects in a similar manner to Equation \eqref{eq:local_retrieval}.
\begin{equation}
\vect{v}_{j}^{m,b,u}
=
\vect{x}^{(m,b)}_{\mathrm{tx}}
+
\text{proj}_{\mathbf{\vect{x}^{(m,u)}_{\mathrm{rx}}-\vect{x}^{(m,b)}_{\mathrm{tx}}}} (\vect{p}^{(m)}_j-\vect{x}^{(m,b)}_{\mathrm{tx}} )
\label{eq:corridor_segment}
\end{equation}
\begin{equation}
d^{\mathrm{cor}}_j
=
\left\|
\vect{p}^{(m)}_j-\vect{v}_{j}^{m,b,u}
\right\|_2.
\label{eq:corridor_distance}
\end{equation}


\paragraph{Sparse multi-scale summaries}
The dense top-$K$ retrieval captures the most relevant individual objects but not the broader spatial distribution of clutter. We therefore add sparse summaries over objects within \(d_{\max}\) that are not necessarily included in the top-$K$ set, computed over increasing radii $\{d_1,\dots,d_R\}$. Around both the transmitter and receiver, we compute object counts and maximum object heights within each radius. Similarly, around the tx-rx corridor, we partition the tx-rx segment into $B$ bins and, within corridor radius $d_{\max}$, compute per-bin summaries of object count and maximum height. These sparse features provide a coarse multi-scale description of environmental density and obstruction patterns. 

Finally, we concatenate the local, dense, and sparse summaries to get our full retrieval context $\vect{x}_m^{tx-rx}$ for a link. This context captures
(i) dense features for the top-$K$ local objects around the receiver and transmitter, (ii) dense features for the top-$K$ objects nearest to the transmitter--receiver corridor, and (iii) sparse multi-scale summaries over local radii and corridor bins. It is both efficient and pair-specific: the dense component preserves the most relevant object-level information, while the sparse component captures broader environmental structure.

\subsection{First-Path Cluster CodeBook}\label{sub_sec:codebook}
Unlike text language modelling, 
the first multipath in the wireless signal is commonly the strongest component, and it often determines the coarse scale of the received channel. Errors at this step propagate to all later decoder states in our autoregressive model, with compounding accuracy effects.
We therefore supplement the first path with a coarse prior that takes advantage of the fact that, for a transmitter, nearby receivers tend to share similar first-path delay and power because they are exposed to similar large-scale geometry, blockage, and LoS or first-reflection structure. Figure~\ref{fig:env_kmeans_overview}{b} illustrates this effect: clustering users by first-path delay and power produces spatially coherent regions. Moreover, the transmitter is often a fixed base station \cite{rappaport2002wireless}, and thus remains consistent between training and test sets for a given environment.

We exploit this regularity with a deliberately small \emph{First-Path Cluster Codebook}. 
The codebook aims to capture a small set of dominant spatial modes while leaving the transformer responsible for the full continuous path prediction problem, including angles, phase, interactions, sequence length, and all subsequent paths. For each transmitter $b$ in environment $m$, we fit $K$-Means on the first-path power and delay of all corresponding receivers in the training set: 
\begin{equation}
\{\vect{c}^{(m,b)}_k\}_{k=1}^{K}
=
\mathrm{KMeans}\!\left(
\{ (\rho_{1}^{(m,b,u)}, \tau_{1}^{(m,b,u)}) \ :u\in\mathcal{U}^{\mathrm{train}}_m\}
\right),
\label{eq:kmeans_centers}
\end{equation}
where $\vect{c}^{(m,b)}_k\in\mathbb{R}^2$ is the $k$-th cluster centroid of the power and delay. We illustrate example clusters in Figure ~\ref{fig:env_kmeans_overview} b). We also compute the within-cluster dispersion of the power and delay for all users within a cluster
\begin{equation}
\vect{\sigma}^{(m,b)}_k
=
\mathrm{Std}\!\left(
\{(\rho_{1}^{(m,b,u)}, \tau_{1}^{(m,b,u)}): z^{(m,b,u)}=k\}
\right),
\label{eq:kmeans_std}
\end{equation}
where $z^{(m,b,u)}$ is the cluster assignment.

At test time, a receiver is assigned a first-path prior through nearest-neighbor transfer within the same transmitter:
\begin{equation}
u^\star
=
\arg\min_{u' \in \mathcal{U}^{\mathrm{train}}_m(b)}
\left\|
\vect{x}^{(m,u)}_{\mathrm{rx}}
-
\vect{x}^{(m,u')}_{\mathrm{rx}}
\right\|_2^2,
\label{eq:nn_assignment}
\end{equation}
and the corresponding cluster prior is
\begin{equation}
\vect{c}_{m,b,u}=\vect{c}^{(m,b)}_{z^{(m,b,u^\star)}},
\qquad
\vect{\sigma}_{m,b,u}=\vect{\sigma}^{(m,b)}_{z^{(m,b,u^\star)}}.
\label{eq:assigned_cluster_prior}
\end{equation}

\textbf{Residual first-path prediction.}
We use the assigned KMeans centroid as a strong initialization for the first path and train the model to predict only the residual correction:
\begin{equation}
\widehat{\vect{y}}^{(1)}_{m,b,u}
=
\vect{c}_{m,b,u}
+
f_{\mathrm{res}}
\Big(
\vect{h}_1,\,
\vect{c}_{m,b,u},\,
\vect{\sigma}_{m,b,u},\,
\mathbf{x}_{m,b,u}
\Big),
\label{eq:first_path_residual}
\end{equation}
where $\vect{h}_1$ is the decoder hidden state at the first generation step and $\mathbf{x}_{m,b,u}$ is the Environmental RAG context. This design separates two sources of structure. The codebook captures the coarse dominant-path mode that is easy to estimate from neighboring training receivers, while the residual network uses the retrieved scene geometry and decoder state to make link-specific corrections. In other words, the $K$-Means prior provides a robust default hypothesis, and Environmental RAG provides the scene evidence needed to refine it. For subsequent paths $t\geq 2$, the model reverts to the standard autoregressive prediction defined by the MultiPathFormer decoder.

\section{Downstream Task Reuse}\label{sub_sec:down}
We demonstrate MultiPathFormer's utility as a wireless foundation model by using it on the tasks of \textit{Sub-6 GHz to mmWave beam prediction, user localization, LoS classification, and channel estimation}. These are commonly used in prior work on wireless foundation models~\cite{yang2025wirelessgpt, alikhani2024lwm, liu2025wifo}, allowing for a direct performance comparison in Section~\ref{sub_sec:exp_down}.

\textbf{Adapter finetuning.} We use adapter finetuning for beam prediction and user localization, and evaluate LoS classification in both zero-shot and adapter-finetuned settings.  For each downstream task, the adapter finetuning leverages the generalized representations from the MultiPathFormer backbone, enabling efficient adaptation through lightweight MLP heads. We apply a pooling function (e.g., first, mean, or max) to the hidden representations produced during autoregressive rollout, as shown in Equation~\eqref{eq:pool}. We freeze the MultiPathFormer backbone and fine-tune a lightweight MLP head $f_{task}$ for downstream tasks as in Equation~\eqref{eq:localization}.
\begin{equation}
\bar{\vect{h}}_{m,n} = \text{Pool}\big(\vect{h}_1,\ldots,\vect{h}_{\hat{L}_{m,n}}\big)
\label{eq:pool}
\end{equation}
\begin{equation}
\hat{\vect{y}}_{task} = f_{task}(\bar{\vect{h}}_{m,n}).
\label{eq:localization}
\end{equation}

 For user localization, the receiver location, RAG, and codebook inputs are zeroed because user location is the prediction target. The multipath sequence input to the decoder produces embeddings for disambiguating the user locations.

For zero-shot LoS, we simply classify based on the dominance of the maximum path power over the other multipaths.

\textbf{Channel estimation.} Similar to other works \cite{liu2025wifo, alikhani2024lwm}, we consider the task of predicting the unknown subcarrier channels from observed noisy subcarriers. MultiPathFormer does not directly compute the channel from predicted multipaths due to the high variance of the phase in high-frequency propagation. Instead, we use MultiPathFormer as a geometry prior to recover the unknown subcarriers from the observed known subcarriers. For each predicted path $\hat{\vect{\pi}}_k$, we define $\vect{b}_k$ as the unit-amplitude, zero-initial-phase channel response induced by that path across antennas and subcarriers. Concretely, for subcarrier $f$,
\[
\vect{b}_k(f)=
e^{-j2\pi f \hat{\tau}_k}
\,\vect{a}_{\mathrm{rx}}(\hat{\theta}^{\mathrm{AoA}}_k)
\,\vect{a}_{\mathrm{tx}}(\hat{\theta}^{\mathrm{AoD}}_k)^{H},
\]
and the MultiPathFormer-derived basis is $\vect{B}=[\vect{b}_1,\ldots,\vect{b}_K]$. Letting $\vect{h}$ denote the channel and $\Omega$ the observed known-subcarrier entries, we estimate the complex coefficients of the predicted path basis by
\begin{equation}
\vect{h} \approx \vect{B}\vect{\alpha},
\qquad
\hat{\vect{\alpha}}=
\arg\min_{\vect{\alpha}}
\left\|\vect{B}_{\Omega}\vect{\alpha}-\vect{h}_{\Omega}\right\|_2^2
+ \lambda \left\|\vect{\alpha}\right\|_2^2 .
\end{equation}
The final estimate is reconstructed as $\vect{B}\hat{\vect{\alpha}}$, so MultiPathFormer provides the sparse propagation support while the observed subcarriers provide the instantaneous complex coefficients needed to infer the unknown subcarriers. This helps the model capture the real-time dynamics of the channel.





\section{Experimental Validation}
We evaluate the proposed MultiPathFormer framework from several perspectives. In Section \ref{subsec:generation_quality}, we assess the contribution of the first-path cluster codebook and environmental retrieval-augmented generation (RAG) to multipath generation quality. In Section \ref{sub_sec:exp_down}, we compare MultiPathFormer against existing state-of-the-art models on a range of downstream tasks. We then evaluate the model's interpretability in Section \ref{sub_sec:interpretability} and ability to generalize to previously unseen user locations and environments in Section \ref{sub_sec:finetune}. Finally, in Section \ref{sub_sec:inference_cost}, we investigate its computational suitability for real-time inference.

\subsection{Experimental Setup}

We evaluate our method using the DeepMIMO dataset framework \cite{DBLP:journals/corr/abs-1902-06435}, which provides reproducible site-specific wireless datasets derived from ray-tracing simulations and is a widely used benchmark in other related works \cite{guler2025robust, alikhani2024lwm}. DeepMIMO is especially well suited to our setting because it exposes not only channel responses but also the geometric and path-level propagation structure underlying those channels. The dataset contains data from different high-fidelity ray-tracing such as  NVIDIA Sionna RT \cite{hoydis2023sionnaopensourcelibrarynextgeneration}, Remcom Wireless InSite \cite{remcom}, and NVIDIA AODT \cite{aodt}. 
We consider a diverse set of  31 environment scenarios for our study. Unless otherwise stated, MultiPathFormer is pretrained on 80\% of the users from 27 scenarios and then evaluated in two zero-shot regimes: (i) the held-out 20\% users from those same 27 scenarios, and (ii) four unseen scenarios. The Appendix gives more details on the MultiPathFormer setup.

\textbf{Baselines and ablations.} Unless otherwise stated, we compare three variants of MultiPathFormer: \emph{Direct}, which autoregressively predicts the path sequence from transmitter and receiver geometry alone; \emph{\(+\)Codebook}, which adds the first-path cluster codebook and residual prediction described in Section~\ref{sub_sec:codebook}; and \emph{\(+\)RAG}, which further adds the Environmental RAG features described in Section~\ref{sub_sec:env_rag}. For the downstream tasks, we use lightweight heads on top of frozen backbone features for beam prediction, user localization, and LoS classification, with zero-shot evaluation considered for channel estimation and LoS classification. All environment codebook clusters are fit using training tx-rx pairs only. 

We compare against LWM \cite{alikhani2024lwm} and WiFo \cite{liu2025wifo}, two commonly used channel-based foundation models, for the downstream tasks; as well as a task-specific beam classification MLP \cite{alrabeiah2020beamblockage} and channel estimation LSTM \cite{channel_lstm}. The MultiPathFormer backbone used in the downstream-task evaluation refers to the raw pretrained foundation model without scenario-specific finetuning, and the reported metrics are computed on held-out test users from each target scenario.

\subsection{Generative Quality and Architectural Ablations}

\label{subsec:generation_quality}

\begin{table*}[t]
\centering
\caption{Cross-scenario path-generation MAE over 31 DeepMIMO scenarios. Delay is reported in microseconds, power in dB, angular errors in degrees, and interaction metrics as multi-label classification scores. Lower is better for all metrics except interaction accuracy/F1, where higher is better.}
\label{tab:main_mae_results}
\setlength{\tabcolsep}{4pt}
\small
\resizebox{\textwidth}{!}{%
\begin{tabular}{lcccccccc}
\toprule
Model & Delay & Power & AoA Az & AoA El & AoD Az & AoD El & Int. Acc. & Int. F1 \\
\midrule
Direct & 0.111 $\pm$ 0.062 & 7.544 $\pm$ 2.475 & 33.106 $\pm$ 11.941 & 4.890 $\pm$ 1.751 & 31.936 $\pm$ 12.001 & 4.042 $\pm$ 1.614 & 0.870 $\pm$ 0.077 & 0.786 $\pm$ 0.147 \\
+Codebook & 0.095 $\pm$ 0.054 & 4.582 $\pm$ 1.851 & 30.097 $\pm$ 10.297 & 4.688 $\pm$ 1.744 & 28.976 $\pm$ 11.066 & 3.778 $\pm$ 1.539 & 0.884 $\pm$ 0.079 & 0.813 $\pm$ 0.145 \\
+RAG & \textbf{0.068} $\pm$ \textbf{0.038} & \textbf{3.081} $\pm$ \textbf{1.383} & \textbf{22.076} $\pm$ \textbf{10.166} & \textbf{3.610} $\pm$ \textbf{1.655} & \textbf{20.147} $\pm$ \textbf{9.357} & \textbf{2.964} $\pm$ \textbf{1.493} & \textbf{0.925} $\pm$ \textbf{0.050} & \textbf{0.890} $\pm$ \textbf{0.079} \\
\bottomrule
\end{tabular}%
}
\end{table*}
 
\begin{figure}[t]
    \includegraphics[height=4cm,width=\linewidth]{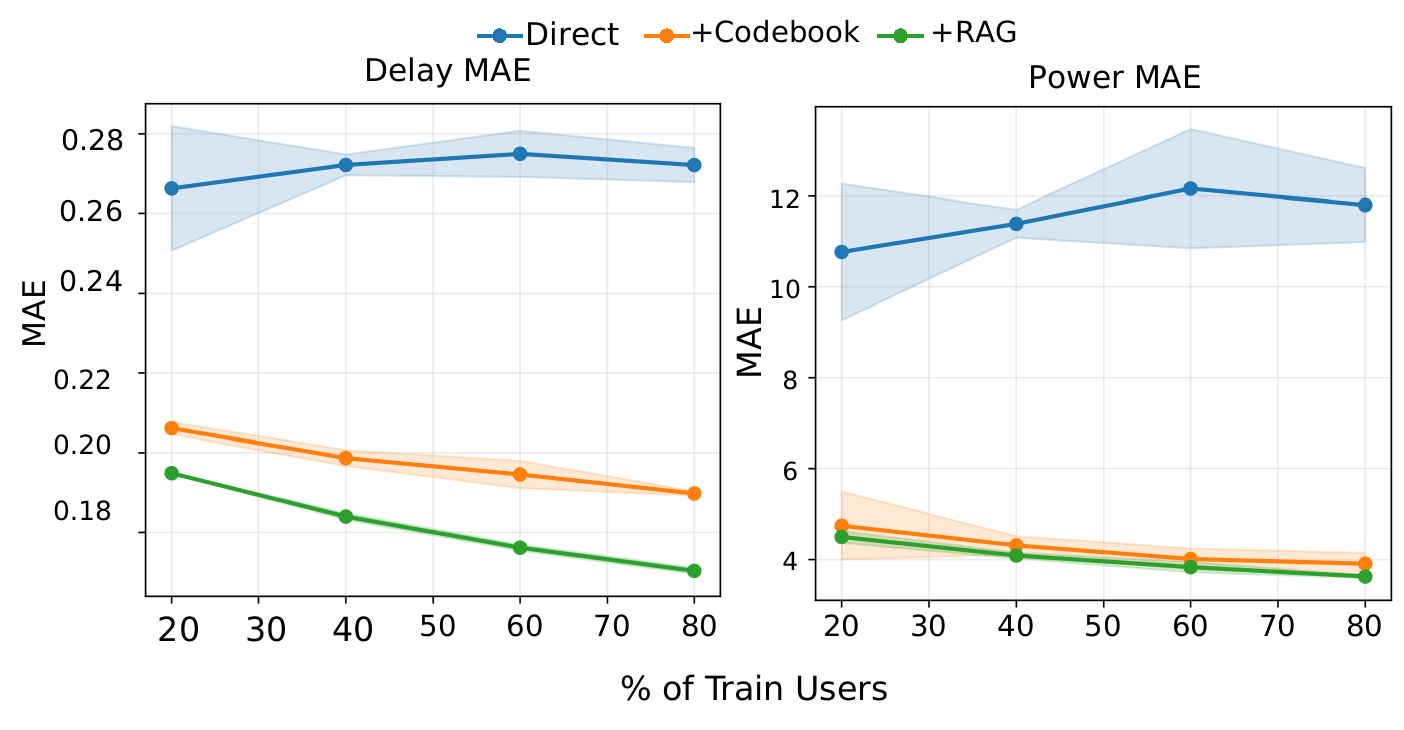}
    \caption{
MultiPathFormer retains low MAE with fewer training users.}
    \label{fig:noise_sensitivity}
\end{figure}

\begin{figure*}[t]
    \centering
    \begin{minipage}[t] {0.4\textwidth}
        \centering
        \includegraphics [height=4cm]{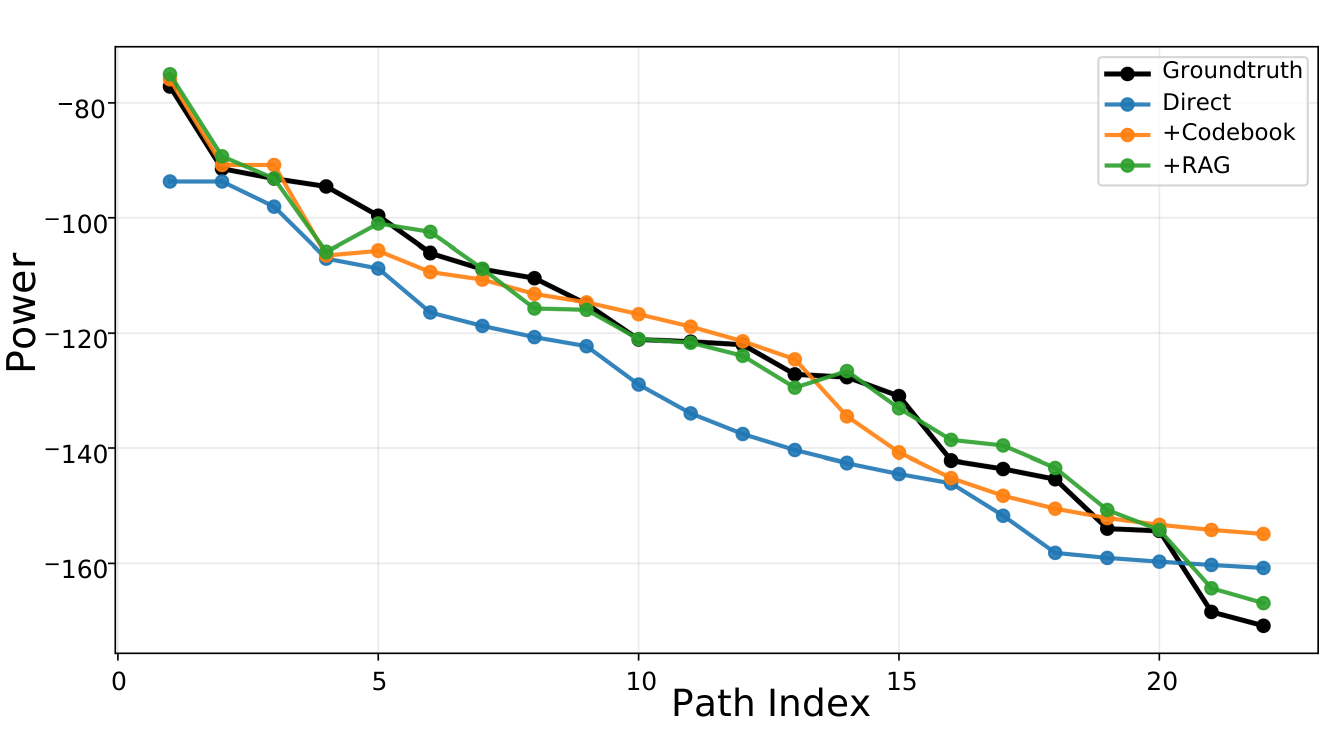}
        \small \textbf{a)} Power versus path index
    \end{minipage}
    \hfill
    \begin{minipage}[t]{0.4\textwidth}
        \centering
        \includegraphics[height=4cm]{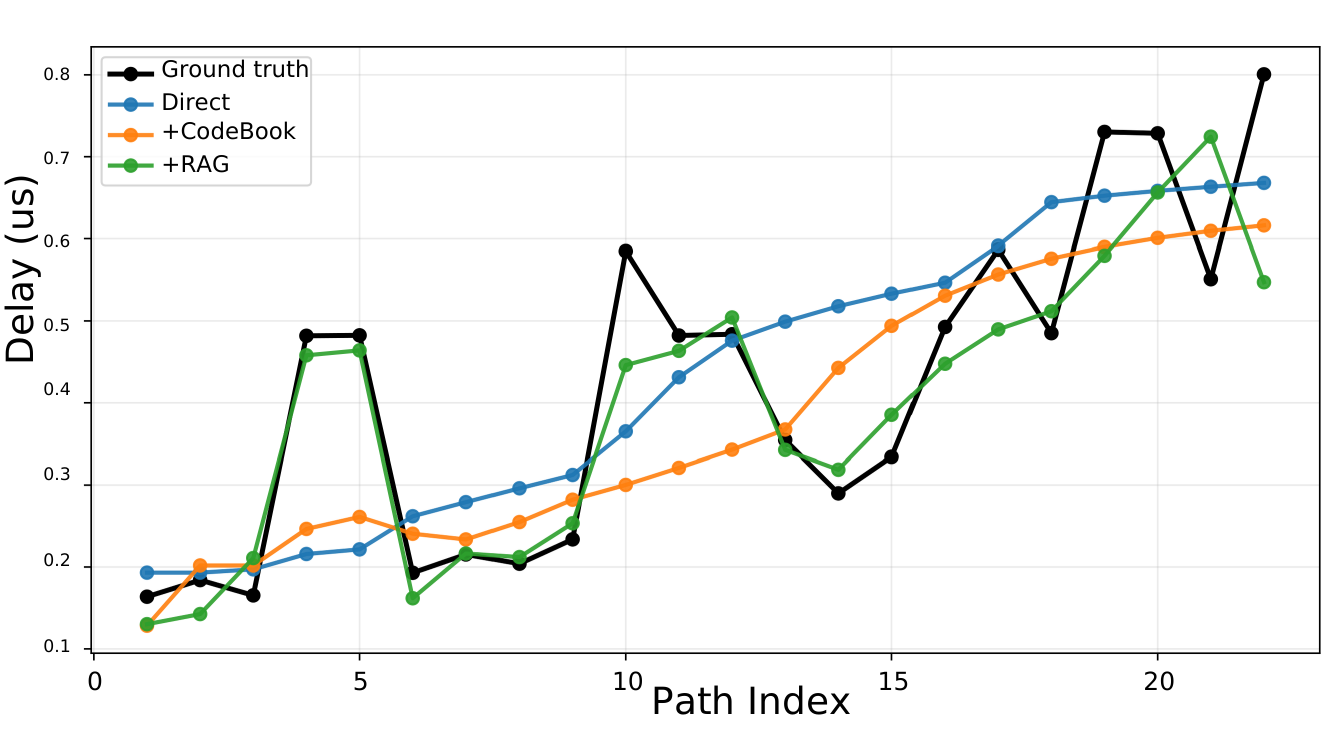}
        \small \textbf{b)} Delay versus path index
    \end{minipage}
    \caption{Qualitative generation example for a receiver in a sample city. Direct captures the broad trend but drifts on early dominant paths. The codebook improves the dominant-path regime, while the full corridor-aware model more closely tracks the ground-truth power decay and delay evolution.}
    \label{fig:houston_generation_qualitative}
\end{figure*}
We train the MultiPathFormer model from scratch for each scenario to test the importance of each of our modules. Table~\ref{tab:main_mae_results} shows that \textit{each structural addition improves the MultiPathFormer backbone}. Adding only the first-path codebook reduces MAE (mean absolute error) of the estimated power from 7.54 to 4.58 (averaged over both held-out users and unseen environments from all scenarios) and improves estimated interaction's F1 from 0.786 to 0.813, showing that a compact dominant-path prior stabilizes the most consequential generation step. Adding Environmental RAG yields the best result on every metric. Relative to the direct MultiPathFormer baseline, the full \(+\)RAG model reduces delay MAE by \(38.7\%\) and power MAE by \(59.2\%\), and improves interaction accuracy from 0.870 to 0.925, and interaction F1 from 0.786 to 0.890. These gains support the main design hypothesis of MultiPathFormer: accurate multipath generation benefits from both a robust first-path prior and pair-specific environmental context. Figure~\ref{fig:noise_sensitivity} shows that this result persists for low-data regimes. Even with only 10\% of users in the training set, the codebook remains beneficial and MAE remains low. 

Figure \ref{fig:houston_generation_qualitative} illustrates the same effect qualitatively. The direct
autoregressive model learns the coarse decay profile but makes
larger errors on the first few paths. The codebook stabilizes
the first path, and the full +RAG model better follows both
the early high-power regime and the later path sequence. 

\subsection{Spatial Coherence and Interpretability}\label{sub_sec:interpretability}

\begin{figure}
    \centering
    \includegraphics[width=\linewidth]{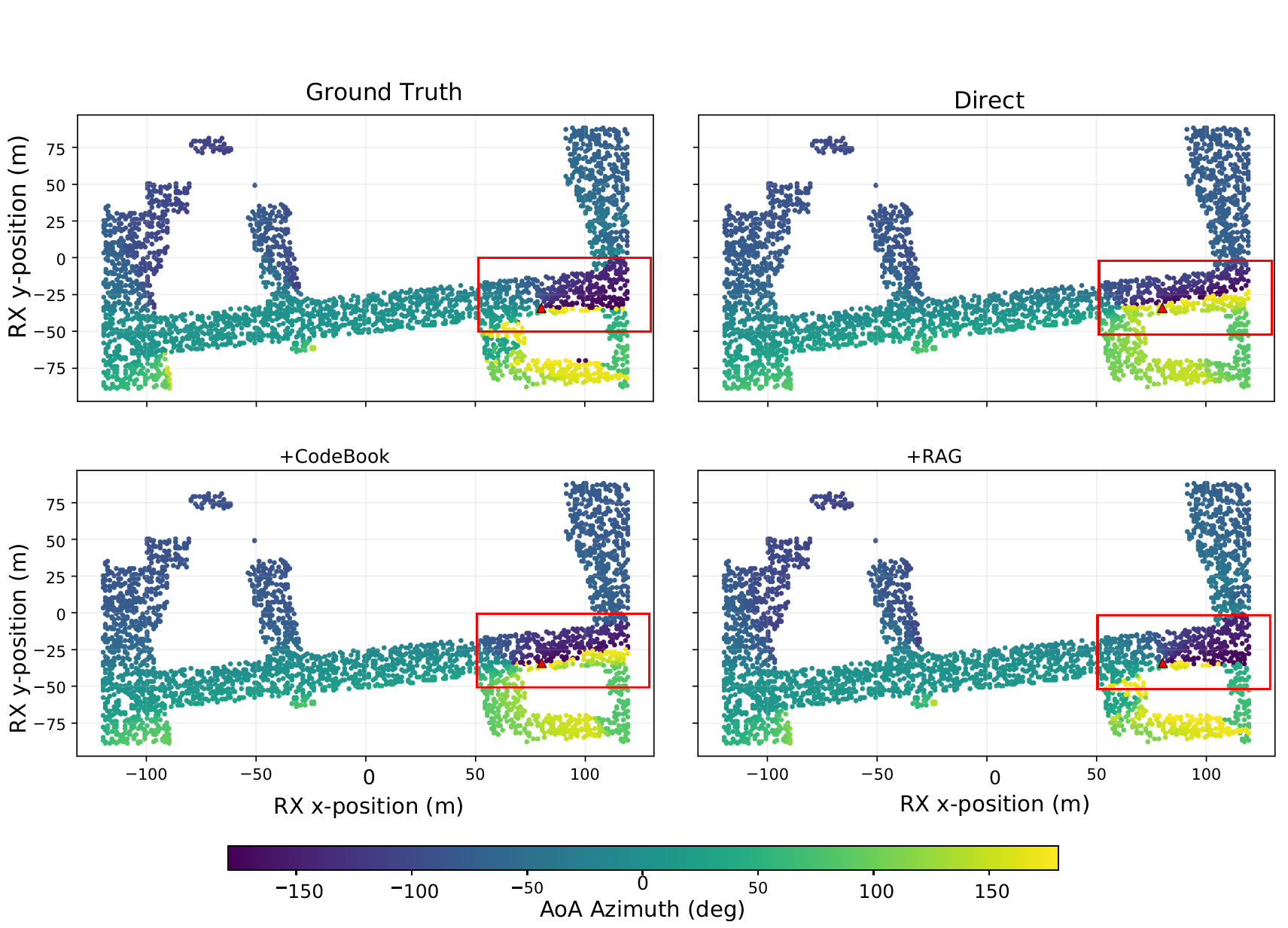}
    \caption{
Spatial comparison of predicted AoA azimuth for one transmitter. The highlighted region in red shows that the full $+RAG$ improves MultiPathFormer's spatial consistency.
    }
    \label{fig:spatial_aoa_coherence}
\end{figure}

We now evaluate MultiPathFormer's interpretability and spatial coherence. Figure~\ref{fig:spatial_aoa_coherence} shows the predicted AoA azimuth field across the user grid. The full model with RAG better matches the spatial structure, showing that retrieval helps align local predictions with the geometry of neighbouring links. This is important because a propagation foundation model should maintain spatial consistency across a deployment in addition to minimizing per-link error. 

\begin{figure}[t]
    \centering
    \includegraphics[height=3.5cm]{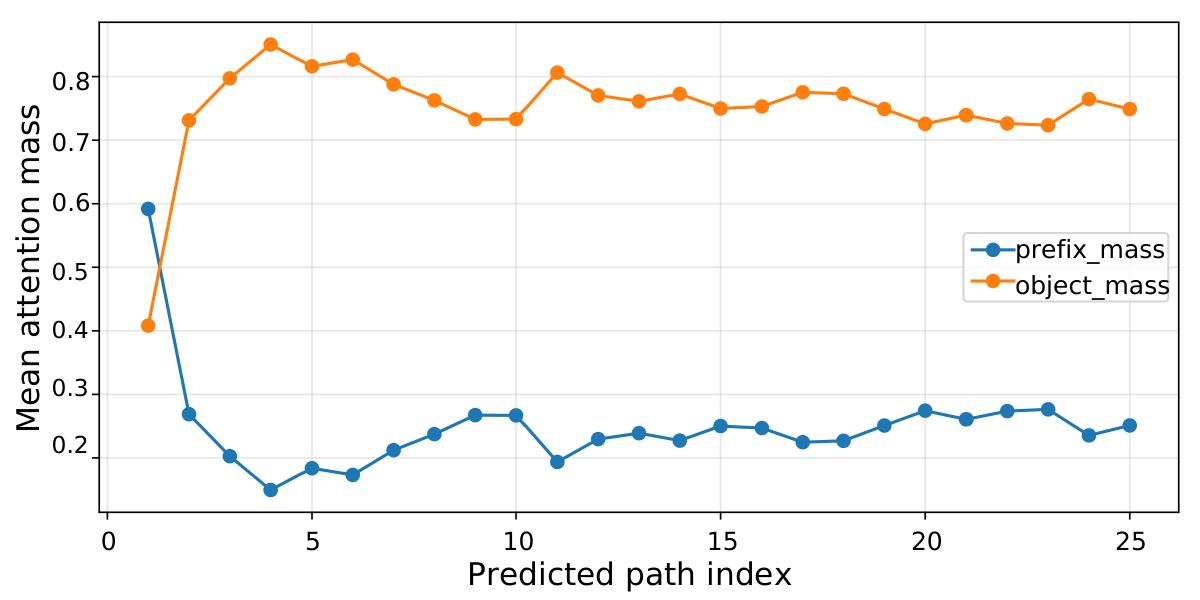}
    \caption{Object-token attention visualization. We expose each retrieved scene object as an individual encoder token and measure decoder cross-attention during teacher-forced rollout. The curve shows the mean attention mass assigned to global prompt tokens versus individual object tokens as a function of generated path index. MultiPathFormer's decoder cross-attends to the prompt more than environment tokens for only the first few paths, consistent with the fact that these are likely LoS.}
    \label{fig:object_token_attention_mass}
\end{figure}

To inspect whether the retrieved environment is used in an interpretable way, we also evaluated an object-token diagnostic variant in which selected scene objects are encoded independently rather than compressed into a small set of aggregate prompt tokens. Fig.~\ref{fig:object_token_attention_mass} shows that the model's decoder cross-attends to the global prompt tokens (tx-rx positions) for the first few generated paths and then attends more to the object tokens for the later paths. This aligns with the physical intuition that the initial paths are mostly LoS and relate more to the positions, compared to later multipaths that interact with a lot of objects in the environments.

\begin{table*}[t]
\centering
\caption{Foundation-model transfer results. The first block reports zero-shot evaluation of the pretrained MultiPathFormer foundation model on held-out users from the 27 pretraining scenarios and on the four unseen scenarios excluded from pretraining. The second block compares scenario-specific adaptation methods averaged over the same 31 scenarios. Lower is better for all metrics except interaction accuracy/F1, where higher is better.}
\label{tab:foundation_transfer}
\setlength{\tabcolsep}{4pt}
\small
\resizebox{\textwidth}{!}{%
\begin{tabular}{lcccccccc}
\toprule
Model / Evaluation Split & Delay & Power & AoA Az & AoA El & AoD Az & AoD El & Int. Acc. & Int. F1 \\
\midrule
\multicolumn{9}{c}{Zero-shot pretrained foundation model} \\
\midrule
Foundation (held-out users in 27 pretraining scenarios) & 0.071 $\pm$ 0.036 & 4.160 $\pm$ 1.818 & 22.753 $\pm$ 9.837 & 3.736 $\pm$ 1.669 & 20.942 $\pm$ 8.840 & 3.108 $\pm$ 1.469 & 0.929 $\pm$ 0.054 & 0.898 $\pm$ 0.082 \\
Foundation (4 unseen scenarios) & 0.326 $\pm$ 0.083 & 14.153 $\pm$ 2.893 & 60.725 $\pm$ 5.987 & 6.545 $\pm$ 1.136 & 61.279 $\pm$ 6.272 & 4.998 $\pm$ 0.533 & 0.716 $\pm$ 0.044 & 0.557 $\pm$ 0.087 \\
\midrule
\multicolumn{9}{c}{Scenario-specific adaptation on the same 31 scenarios} \\
\midrule
Scenario-trained $+RAG$ & 0.068 $\pm$ 0.038 & 3.081 $\pm$ 1.383 & 22.076 $\pm$ 10.166 & 3.610 $\pm$ 1.655 & 20.147 $\pm$ 9.357 & 2.964 $\pm$ 1.493 & 0.925 $\pm$ 0.050 & 0.890 $\pm$ 0.079 \\
Foundation $+$ finetune & \textbf{0.065} $\pm$ \textbf{0.037} & \textbf{2.888} $\pm$ \textbf{1.471} & \textbf{20.208} $\pm$ \textbf{9.570} & \textbf{3.311} $\pm$ \textbf{1.633} & \textbf{18.056} $\pm$ \textbf{8.450} & \textbf{2.654} $\pm$ \textbf{1.389} & \textbf{0.932} $\pm$ \textbf{0.050} & \textbf{0.900} $\pm$ \textbf{0.078} \\
\bottomrule
\end{tabular}%
}
\end{table*}
\subsection{Cross User-Environment Transfer}\label{sub_sec:finetune}
We next demonstrate MultiPathFormer's generalizability in zero-shot transfer to new user locations and for scenario-level adaptation.
Table~\ref{tab:foundation_transfer} evaluates how the foundation backbone transfers to unseen user locations and environments. In the zero-shot setting, MultiPathFormer transfers well to held-out users within the 27 pretraining environments, achieving average MAEs of 0.071 delay and 4.160 power, with 0.898 interaction F1. Performance drops on the four fully unseen scenarios, indicating that zero-shot cross-environment transfer remains challenging when the target geometry is absent from pretraining. However, finetuning the foundation checkpoint on each target scenario provides a stronger initialization than training the same $+RAG$ architecture from scratch, improving the 31-scenario average delay MAE from 0.068 to 0.065 and power MAE from 3.081 to 2.888. 


\subsection{Downstream Tasks}\label{sub_sec:exp_down}
We next compare MultiPathFormer's performance on Section~\ref{sub_sec:down}'s four downstream tasks to demonstrate the superiority of our path-based model architecture compared to prior channel-based architectures. While we use the full MultiPathFormer for this comparison, which utilizes environment and transmitter knowledge that standard channel models do not, based on Table~\ref{tab:main_mae_results}'s results  MultiPathFormer's simpler versions without this information would still achieve at least comparable performance to the channel-based baselines.
\subsubsection{Beam Prediction}
We evaluated whether models can use sub-6 GHz channels/multipaths for mmWave beam selection similar to the setup in \cite{alrabeiah2020beamblockage, alikhani2024lwm}.
For the downstream tasks below, we report averages over the scenarios shared by the compared baselines for each task. The MultiPathFormer foundation model achieves an average top-1 accuracy of 0.671 and top-3 accuracy of 0.914, outperforming the MLP, WiFo, and LWM baselines. This result is consistent with the intuition that the optimal mmWave beam direction is determined primarily by the dominant path's angle of departure, which is explicitly represented in MultiPathFormer's path token sequence. Channel-level models must implicitly recover this geometric structure from the collapsed channel tensor, while MultiPathFormer has direct access to it through the multipath representation.

\begin{table*}[t]
\centering
\caption{Average downstream-task performance averaged over the scenarios. MultiPathFormer outperforms LWM and WiFo.}
\label{tab:downstream_results}
\setlength{\tabcolsep}{3pt}
\footnotesize
\resizebox{\textwidth}{!}{%
\begin{tabular}{lccccccccc}
\toprule
& \multicolumn{2}{c}{Beam} & \multicolumn{3}{c}{Localization} & \multicolumn{2}{c}{LoS} & \multicolumn{2}{c}{Channel Est.} \\
\cmidrule(lr){2-3} \cmidrule(lr){4-6} \cmidrule(lr){7-8} \cmidrule(lr){9-10}
Model & Top-1 & Top-3 & Mean (m) & Median (m) & P90 (m) & Acc. & F1 & NMSE & NMSE (dB) \\
\midrule
MLP & 0.450 $\pm$ 0.054 & 0.672 $\pm$ 0.087 & 82.25 $\pm$ 17.30 & 80.82 $\pm$ 21.59 & 131.57 $\pm$ 22.46 & 0.936 $\pm$ 0.092 & 0.884 $\pm$ 0.263 & -- & -- \\
LSTM & -- & -- & -- & -- & -- & -- & -- & \textbf{0.049 $\pm$ 0.019} & \textbf{-14.78 $\pm$ 1.66} \\
WiFo & 0.369 $\pm$ 0.057 & 0.616 $\pm$ 0.081 & 68.79 $\pm$ 9.08 & 58.52 $\pm$ 9.52 & 136.00 $\pm$ 22.61 & 0.962 $\pm$ 0.028 & 0.946 $\pm$ 0.040 & 1.321 $\pm$ 0.404 & 0.93 $\pm$ 1.56 \\
LWM & 0.429 $\pm$ 0.037 & 0.648 $\pm$ 0.053 & 72.78 $\pm$ 13.30 & 69.86 $\pm$ 15.08 & 123.60 $\pm$ 21.73 & 0.978 $\pm$ 0.018 & 0.968 $\pm$ 0.028 & 1.000 $\pm$ 0.000 & 0.00 $\pm$ 0.00 \\
MultiPathFormer & \textbf{0.671 $\pm$ 0.052} & \textbf{0.914 $\pm$ 0.030} & \textbf{5.57 $\pm$ 2.01} & \textbf{4.15 $\pm$ 1.27} & \textbf{10.13 $\pm$ 3.09} & \textbf{0.994 $\pm$ 0.003} & \textbf{0.990 $\pm$ 0.008} & 0.561 $\pm$ 0.353 & -13.27 $\pm$ 4.40 \\
\bottomrule
\end{tabular}%
}
\end{table*}


\subsubsection{User Localization}




Table~\ref{tab:downstream_results} also evaluates whether the learned path-level representation supports inverse geometric inference. Because receiver coordinates are unknown by definition during localization, we zero all codebook and RAG
prompt slots that depend on receiver location when extracting
localization features. MultiPathFormer performs best on all three localization metrics, with a mean error of 5.57 m, median error of 4.15 m, and 90th-percentile error of 10.13 m over the common scenarios. This suggests that the
MultiPathFormer decoder encodes useful geometric information
from the generated multipath structure itself and is not simply exploiting receiver-dependent codebook or RAG features.

\subsubsection{LoS Classification}

We first evaluate MultiPathFormer zero-shot for LoS classification by measuring the dominance of the maximum-power path over the remaining paths. This zero-shot rule achieves 0.976 accuracy and 0.959 F1, showing that the LoS/NLoS structure is already captured by the generated multipath. Table~\ref{tab:downstream_results} shows that adapter finetuning of the same pretrained backbone improves further to 0.994 accuracy and 0.990 F1, outperforming the other methods.

\subsubsection{Channel Estimation}
We finally compare the MultiPathFormer channel-estimation reuse strategy described in Section~\ref{sub_sec:down} against the channel-level LSTM, WiFo, and LWM baselines on the common evaluation scenarios. The LSTM is trained from scratch on each target scenario, while MultiPathFormer, WiFo, and LWM are evaluated without scenario-specific channel-estimation finetuning. The antenna setup is a \(1\times 8\) array with 32 subcarriers. For all methods, the first half of the subcarriers are observed and the remaining half are estimated. Table~\ref{tab:downstream_results} shows that the scenario-trained LSTM achieves the lowest NMSE, while MultiPathFormer is the strongest non-finetuned foundation-model baseline.

\subsection{Inference Cost}\label{sub_sec:inference_cost}

\begin{table}[t]
\centering
\caption{All three MultiPathFormer configurations maintain low inference times (ms), averaged across different scenarios.}\label{tab:timing_results}
\small
\begin{tabular}{lcccc}
\toprule
Model & Params (M) & Infer & CodeBook & RAG \\ 
\midrule
Direct & 35.256 & 2.77 & -- & -- \\
\(+\) Codebook & 35.261 & 3.39 & 0.92 & -- \\
\(+\) RAG & 35.512 & 3.45 & 0.92 & 0.21 \\
\bottomrule
\end{tabular}
\end{table}
Finally, we show that \textit{MultiPathFormer enables fast inference}. Table~\ref{tab:timing_results} shows that all three variants remain in the same low-latency regime, between 2--3 ms per inference. The direct model is fastest because it avoids codebook and RAG lookup, while the full \(+\)RAG model adds only modest overhead and a small parameter increase (35.5M vs. 35.26M) while delivering substantially better generation fidelity. This makes the full MultiPathFormer design attractive not only in terms of accuracy but also in terms of deployment practicality.




\section{Conclusion and Future Direction}

In this paper, we presented MultiPathFormer, a wireless foundation model that treats multipath propagation, rather than the channel tensor, as the pretraining object. By representing each transmitter--receiver link as an autoregressive sequence of path tokens, PathFormer learns reusable path-level representations that preserve dominant-path structure, angular geometry, and propagation interactions. An important next step is to evaluate whether these gains carry over from ray-tracing data to actual real-world measurement scenarios.
\appendix
\section{Additional Model and Training Details}
\label{appendix:model_training}


\paragraph{Model Architecture} MultiPathFormer uses an encoder-decoder architecture with a full transformer decoder and an MLP encoder. The model uses 12 transformer layers with 8 heads and a hidden dimension of 1024, supporting a maximum sequence length of 25 paths.  
\paragraph{RAG and Corridor retrieval configuration}
The first-path codebook uses \(K=25\) KMeans clusters per transmitter. For the environmental retrieval module, we use \(K_{\mathrm{local}}=5\) nearest local objects, \(K_{\mathrm{corr}}=5\) nearest corridor objects, \(B=8\) corridor bins, and local/corridor radii \(\{25, 50, 100\}\) meters. Each dense object descriptor includes distance, height, footprint area, and material properties (permittivity, conductivity, and scattering coefficient) when available.

\paragraph{Training setup}
We split users within each transmitter into train and validation sets and retain only users with at least one valid path. Path delays are converted to microseconds, powers are scaled by \(0.01\), and angles are represented in radians. All models are trained with batch size \(128\) using AdamW and cosine-annealing warm restarts. We train the foundation model for \(200\) epochs with learning rate \(2\times10^{-5}\), $\lambda_{len}$ and $\lambda_{int}$  weights to \(0.01\). During training, we randomly add noise to 20\% of the tokens in every path during every epoch. This served as a data augmentation strategy to enable the model to recover from generation errors.

\paragraph{Evaluation protocol}
The model is evaluated in full autoregressive generation during inference with no teacher-forcing. The results are for each path averaged over the actual number of multipaths and then over the users.

\paragraph{Environment Summary}
The main generation study uses 31 DeepMIMO scenarios. Across these environments, the number of active users ranges from \(25{,}669\) to \(320{,}104\), the LoS user ratio ranges from \(0.042\) to \(0.763\), the mean number of paths per active user ranges from \(2.3\) to \(21.4\), and the mean transmitter--receiver distance ranges from \(77.8\) m to \(178.7\) m. This diversity covers both dense urban scenes with many buildings and more open environments with fewer obstructions, providing a broad testbed for evaluation.

\bibliographystyle{IEEEtran}
\bibliography{references}

\end{document}